\PassOptionsToPackage{table,xcdraw}{xcolor}  
\documentclass[sigconf,screen,nonacm]{acmart}

\usepackage{xcolor}  

\usepackage{graphicx}
\usepackage{booktabs}
\usepackage{multirow}

\definecolor{lightgreen}{rgb}{0.6, 0.9, 0.6}
\definecolor{lightyellow}{rgb}{1.0, 1.0, 0.6}
\definecolor{lightorange}{rgb}{1.0, 0.8, 0.6}
\definecolor{lightpurple}{rgb}{0.8, 0.7, 1.0}
\definecolor{lightpink}{rgb}{1.0, 0.8, 0.9}
\definecolor{lightgray}{rgb}{0.9, 0.9, 0.9}
\definecolor{darkred}{rgb}{0.65, 0.0, 0.0}
\definecolor{darkblue}{rgb}{0, 0.4, 0.75}

\usepackage[linesnumbered,ruled,vlined]{algorithm2e}

\usepackage{wrapfig}
\usepackage{tikz}
\usepackage{comment}

\usepackage{amsmath,amssymb}
\usepackage{color}
\usepackage{paralist}
\usepackage{xcolor}
\usepackage{booktabs}  
\usepackage{multirow} 
\usepackage{multicol} 
\usepackage{arydshln}
\usepackage{verbatim}
\usepackage{arydshln}
\usepackage{mathtools}
\usepackage{verbatim}
\usepackage{enumitem}
\usepackage{enumerate}
\usepackage{makecell}
\usepackage{comment}
\usepackage{paralist}
\usepackage{bbding}
\usepackage{cleveref}
\usepackage{tabularx}

\begin{document}

\title{Self-Controlled Dynamic Expansion Model for Continual Learning}

\author{Runqing Wu}
\affiliation{
  \institution{Huazhong University of Science and Technology}
  \city{Wuhan}
  \country{China}
}

\author{Kaihui Huang}
\affiliation{
  \institution{University of Electronic Science and Technology of China}
  \city{Shenzhen}
  \country{China}
}

\author{Hanyi Zhang}
\affiliation{
  \institution{Technische Universität München}
  \city{München}
  \country{Germany}
}

\author{Fei Ye}
\authornote{Corresponding author.}
\affiliation{
  \institution{University of Electronic Science and Technology}
  \city{Chengdu}
  \country{China}
}

\begin{abstract}
Continual Learning (CL) epitomizes an advanced training paradigm wherein prior data samples remain inaccessible during the acquisition of new tasks. Numerous investigations have delved into leveraging a pre-trained Vision Transformer (ViT) to enhance model efficacy in continual learning. Nonetheless, these approaches typically utilize a singular, static backbone, which inadequately adapts to novel tasks, particularly when engaging with diverse data domains, due to a substantial number of inactive parameters. This paper addresses this limitation by introducing an innovative Self-Controlled Dynamic Expansion Model (SCDEM), which orchestrates multiple distinct trainable pre-trained ViT backbones to furnish diverse and semantically enriched representations. Specifically, by employing the multi-backbone architecture as a shared module, the proposed SCDEM dynamically generates a new expert with minimal parameters to accommodate a new task. A novel Collaborative Optimization Mechanism (COM) is introduced to synergistically optimize multiple backbones by harnessing prediction signals from historical experts, thereby facilitating new task learning without erasing previously acquired knowledge. Additionally, a novel Feature Distribution Consistency (FDC) approach is proposed to align semantic similarity between previously and currently learned representations through an optimal transport distance-based mechanism, effectively mitigating negative knowledge transfer effects. Furthermore, to alleviate over-regularization challenges, this paper presents a novel Dynamic Layer-Wise Feature Attention Mechanism (DLWFAM) to autonomously determine the penalization intensity on each trainable representation layer. An extensive series of experiments have been conducted to evaluate the proposed methodology's efficacy, with empirical results corroborating that the approach attains state-of-the-art performance.
\end{abstract}


\maketitle

\section{Introduction}

The goal of continual learning (CL), also known as lifelong learning, is to create a model that can continuously learn new information while remembering what has already been learned \cite{LifeLong_review}. However, current deep learning models often suffer significant performance degradation in continual learning, mainly from catastrophic forgetting \cite{LifeLong_review}, as these models do not have the mechanisms to prevent information loss when adjusting to new tasks. Because of these benefits, continual learning has been applied to real-world applications in a variety of domains, such as autonomous driving, robotic navigation, and medical diagnostics.

Numerous methods have been developed to solve the problem of network forgetting in the continual learning scenario. These fall into three main categories: the rehearsal-based methods, which optimize a small memory buffer to preserve many important examples \cite{TinyLifelong,RainbowMemory}, the dynamic expansion frameworks, which allow for the automatic construction and integration of new hidden layers and nodes into an existing backbone to capture new information \cite{Adanet,ompactingPicking}; and the regularization-based methods, which add a regularization term to the primary objective function to minimize significant changes to many previously important network parameters \cite{MeasureForgetting,OptimizingNeural}. These methods, however, are primarily focused on addressing catastrophic forgetting while ignoring plasticity which is the ability of learning new tasks. 

In continual learning, achieving an equilibrium between network forgetting and plasticity is paramount to ensuring optimal performance across both historical and current tasks (refer to \cite{NewSigntsCL}). Numerous investigations have advocated for the utilization of the pre-trained Vision Transformer (ViT) \cite{Vit} as a means to mitigate network forgetting while enhancing plasticity \cite{RevisitingCL,Vit,Ranpac}. The semantically enriched representations generated by the pre-trained ViT backbone facilitate rapid adaptation to novel task learning. Nevertheless, these approaches typically rely on a singular pre-trained ViT as the backbone, which may exhibit constrained learning capabilities when confronted with tasks containing information divergent from the pre-trained ViT's stored knowledge. Furthermore, these methodologies often immobilize the parameters of the pre-trained backbone to prevent forgetting, thereby impacting plasticity. This paper introduces a novel framework, the Self-Controlled Dynamic Expansion Model (SCDEM), which concurrently addresses network forgetting and plasticity by managing and optimizing a series of diverse pre-trained ViT backbones to deliver semantically rich representations. By utilizing these backbones as the shared module, a new expert network is dynamically constructed with minimal parameters, aiming to capture information from new task learning. In contrast to existing pre-trained methodologies that employ a single backbone and consequently fail to achieve optimal performance across various specific tasks \cite{RevisitingCL,Vit,Ranpac}, the proposed SCDEM demonstrates robust generalization across diverse data domains.

To augment plasticity within the realm of continual learning, we propose an innovative Collaborative Optimization Mechanism (COM) designed to iteratively refine the backbones, thereby yielding adaptive and resilient representations. In addition, the proposed COM targets the optimization of the last few representation layers of each backbone, thereby mitigating substantial computational demands. To circumvent the issue of negative knowledge transfer, it is imperative that optimizing the backbones should not alter the pre-established prediction patterns of historical experts. To achieve this, the proposed COM freezes and copies the trainable parameters of each backbone as the frozen backbone, aiming to preserve the previously learned representation information on the most recent task. Subsequently, the proposed COM endeavors to minimize the Kullback–Leibler (KL) divergence between predictions derived from both previously and currently acquired backbones, facilitating the incremental assimilation of new information while retaining all previously acquired knowledge.

To further mitigate the adverse effects of negative knowledge transfer, we introduce an innovative Feature Distribution Consistency (FDC) method designed to stabilize the trainable representation layers within neural network backbones during the optimization process. The proposed FDC method conceptualizes the representations derived from multi-level feature layers as feature distributions and seeks to minimize the optimal transport distance between previously acquired and newly learned feature distributions. This strategy ensures the retention of robust, previously acquired representations while facilitating the learning of new tasks. Additionally, to address over-regularization challenges that impede model plasticity, we propose a novel Dynamic Layer-Wise Feature Attention Mechanism (DLFAM). This mechanism manages and optimizes a parametric function to autonomously assess the significance of each representation layer during the regularization process. The proposed DLFAM synthesizes weighted layer-wise features from each backbone into a cohesive representation, forming an augmented feature distribution. An optimal transport distance metric is applied to the augmented feature distributions to guide the model's optimization process, thereby selectively penalizing alterations in each trainable representation layer and circumventing over-regularization issues. A thorough array of experiments centred on continual learning has been executed, illustrating that our proposed methodology markedly exceeds current baselines across all experimental setups. The principal contributions of this research are delineated as follows~:
\begin{itemize}[leftmargin=13pt]
\setlength{\itemsep}{2pt}
\setlength{\parsep}{2pt}
\setlength{\parskip}{2pt}
\item[\large$\bullet$] This paper proposes a novel Self-Controlled Dynamic Expansion Model (SCDEM) that optimizes and manages several different pre-trained ViT backbones to provide semantically rich representations, enhancing the model's performance in cross-domain continual learning.
\item[\large$\bullet$] We propose a novel COM to collaboratively optimize each backbone to adapt to new tasks without forgetting all previously learnt knowledge.
\item[\large$\bullet$] We propose a novel FDC approach to align the semantic similarity between the previously and currently learnt representations, which can minimize the negative knowledge transfer effects.

\item[\large$\bullet$] We propose a novel DLWFAM to automatically determine the importance of each trainable representation layer during the model's regularization process, which can effectively avoid over-regularization issues.

\end{itemize}
\section{Relate Work}
\noindent \textbf{Rehearsal-based methods} remain one of the most fundamental and widely used strategies in continual learning to address the problem of catastrophic forgetting \cite{CLBlurry}. These methods mitigate forgetting by storing a representative subset of previously seen samples and replaying them during the training of new tasks \cite{CLBlurry,Co2l,NoSelection,CLMutual,LearnAdd,LifeLong_combination,OnlineStructuredLaplace,GCR,DualPrototypeCL}. The effectiveness of such methods is highly dependent on the quality of the sample selection. To further enhance performance, rehearsal strategies are often combined with regularization-based approaches through the use of memory buffers \cite{plasticityCL,OptimizingNeural,TinyLifelong,GradientEpisodic,KernelCL,CLBit,CLNullSpace,VCL,Uncertainty_CL,FlattingCL,NPCL}. As an alternative to storing raw data, generative replay methods employ models such as Variational Autoencoders (VAEs) \cite{VAE} and Generative Adversarial Networks (GANs) \cite{GAN} to synthesize previous data distributions \cite{Lifelong_VAE,GenerativeLifelong,Generative_replay,LifelongGAN,Sddgr}, thereby addressing privacy concerns associated with direct data storage.

\noindent 
\textbf{Knowledge distillation (KD)} has also been widely adopted in continual learning, originally developed to transfer knowledge from a larger teacher model to a more compact student model \cite{KD_Review,Distilling_nets}. In the continual learning setting, KD is adapted by treating the model trained on previous tasks as the teacher and the current model as the student. By minimizing the discrepancy between their outputs, the student is guided to retain knowledge from past tasks \cite{Lwf}. Several approaches integrate KD with rehearsal mechanisms into unified frameworks to further improve performance. A notable example is iCaRL \cite{icarl}, which combines rehearsal with a nearest-mean-of-exemplars classifier, enhancing robustness to representation drift. Additionally, self-distillation techniques have been proposed to preserve learned features without relying on external teacher models, effectively alleviating forgetting \cite{Co2l}.

\noindent
\textbf{Dynamic expansion architectures} offers a complementary strategy to fixed-capacity models. While rehearsal and KD-based methods have shown promising results, they often struggle with long task sequences or highly heterogeneous domains. To address this, dynamic and expandable architectures have been proposed, which progressively allocate new sub-networks or hidden layers for incoming tasks, while keeping previously learned parameters frozen to preserve prior knowledge \cite{Adanet,ompactingPicking,LearnAdd,ProgressiveNN,BatchEnsemble,OnlineLearning,ForgetFree,EffecientFeature}. Such approaches allow continual models to scale with task complexity and maintain performance across all learned tasks. More recently, Vision Transformers (ViT) \cite{Vit} have been adopted as modular backbones in dynamic architectures, demonstrating improved scalability and adaptability compared to CNN-based variants \cite{MetaVitCL,Dytox}.

\noindent For a more comprehensive overview of related techniques and comparisons, please refer to the extended discussion in \textbf{Appendix A} from Supplementary Material (SM).

\section{Methodology}

\begin{figure*}[t]
  \centering
  \includegraphics[page=1, width=\textwidth, trim=60 175 70 80, clip]{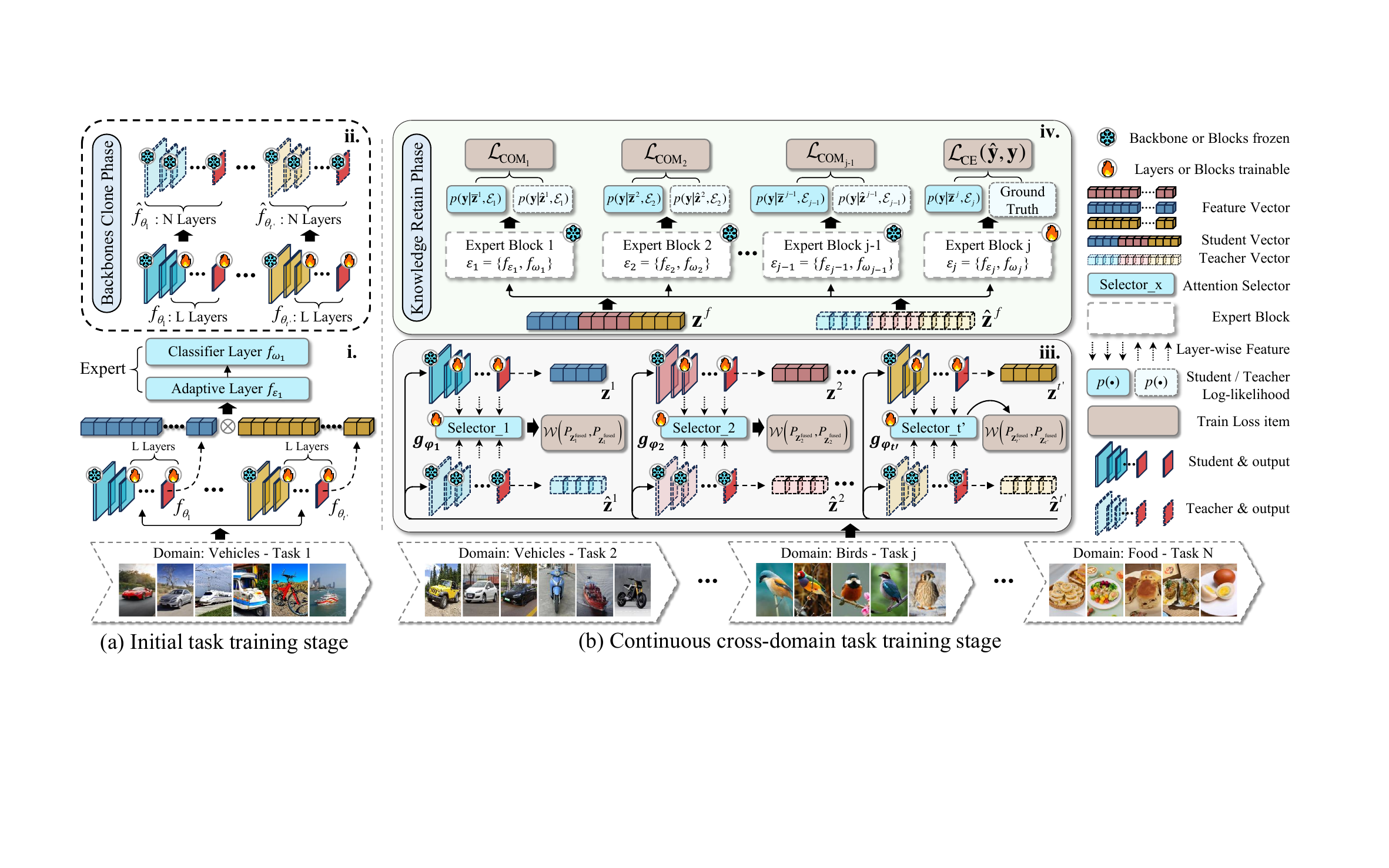}
    \captionsetup{skip=4pt} 
    \caption{
    Overview of the SCDEM training framework. 
    (a) Initial task stage: (i) Each backbone $f_{\theta_j}$ is partially fine-tuned to extract multi-source features ${\bf z}^f$, which are used to train a task-specific expert $\mathcal{E}_t = \{f_{\xi_t}, f_{\omega_t}\}$. 
    (ii) Backbone copies $\hat{f}_{\theta_j}$ are frozen to retain prior knowledge. 
    (b) Continual learning stage: (iii) A selector $g_{\phi_t}$ assigns layer-wise weights to compute ${\bf Z}^{\text{fused}}_j$, aligned with its frozen counterpart via Wasserstein distance. 
    (iv) Knowledge consistency is enforced through KL divergence between expert outputs ($\mathcal{L}_{\rm COM}$), and task-specific supervision is applied via cross-entropy loss ($\mathcal{L}_{\rm CE}$).
    }

  \label{fig1}
  \vspace{-5pt}
\end{figure*}

\subsection{Problem Statement}

In continual learning, a model is trained in a dynamic and non-stationary environment where data arrives sequentially in the form of tasks. At each stage, the model is only allowed to access the training data from the current task, and data from previous tasks is no longer accessible. Let the $i$-th training task be denoted as $D^s_i = { ({\bf x}^i_j, {\bf y}^i_j) }_{j=1}^{n^i}$, and the corresponding test set be $D^t_i = { ({\bf x}^{t,i}_j, {\bf y}^{t,i}j) }_{j=1}^{n^{t,i}}$, where $n^i$ and $n^{t,i}$ represent the number of training and testing samples, respectively. Here, ${\bf x}^{t,i}_j \in \mathcal{X} \subseteq \mathbb{R}^{d_x}$ is the input feature and ${\bf y}^{t,i}_j \in \mathcal{Y} \subseteq \mathbb{R}^{d_y}$ is the corresponding label, with $\mathcal{X}$ and $\mathcal{Y}$ denoting the input and label spaces. In a class-incremental setting, each training dataset $D^s_i$ is partitioned into $C_i$ disjoint subsets: $\{{ D^s_i(1), \cdots, D^s_i(C_i) }\}$, where each subset contains samples belonging to a single or a small group of consecutive classes. Let $\{{ T_1, \ldots, T_{C_i} }\}$ denote the sequence of tasks, with task $T_j$ corresponding to subset $D^s_i(j)$. During training on task $T_j$, the model is restricted to accessing only $D^s_i(j)$, and all previous subsets $\{{ D^s_i(1), \ldots, D^s_i(j-1) }\}$ remain unavailable.

While most existing continual learning approaches focus on learning new categories within a single domain, real-world applications often involve domain heterogeneity. Suppose we are given $t$ domains $\{{ D^s_1, \ldots, D^s_t }\}$, where each $D^s_i$ is further divided into $C_i$ subsets as described above. A sequential data stream $S$ can be defined as: \begin{equation} S = { D^s_1(1), \ldots, D^s_1(C_1), \ldots, D^s_t(C_t) } ,. \end{equation} This scenario introduces challenges from both class-incremental learning and domain shift. After the model finishes training over the entire stream, it is evaluated on the corresponding test sets $\{{ D^t_1, \ldots, D^t_t }\}$ to assess its ability to retain knowledge and generalize across tasks and domains.

\subsection{Framework Overview}

\begin{figure*}[t]
  \centering
  \includegraphics[page=2, width=\textwidth, trim=60 380 60 80, clip]{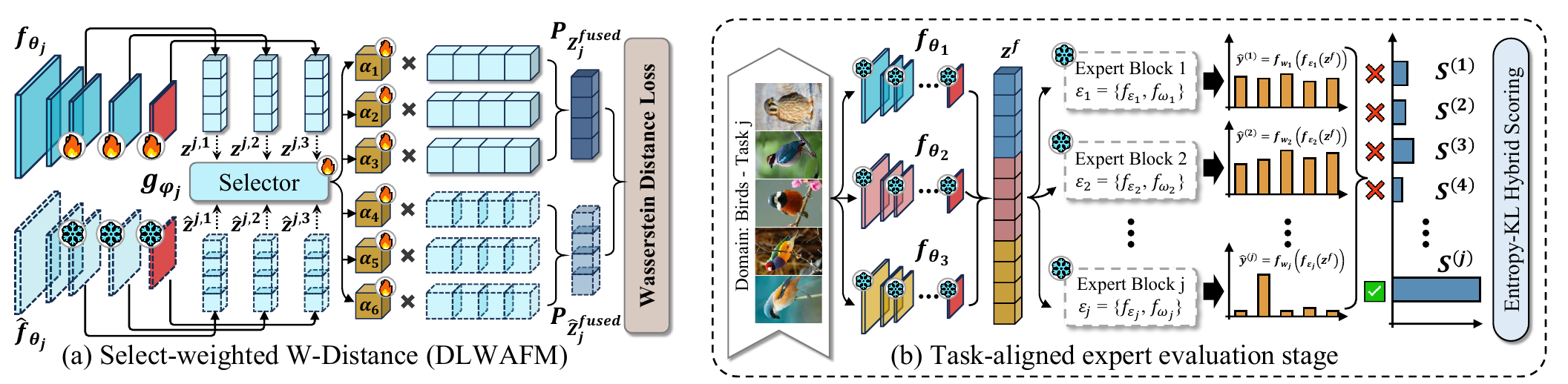}
    \captionsetup{skip=4pt} 
    \caption{
    (a) Selector-weighted fusion(DLWFAM): layer-wise features from $f_{\theta_j}$ are aggregated via attention weights $\{\alpha_k\}$ to form ${\bf Z}^{\text{fused}}_j$, aligned with the frozen $\hat{\bf Z}^{\text{fused}}_j$ via Wasserstein distance. 
    (b) Task-free expert selection: each expert is scored by combining prediction entropy and KL divergence between its log-likelihood and a global softmax distribution, enabling class-IL inference without task labels.
    }

  \label{fig2}
  \vspace{-8pt}
\end{figure*}

\begin{algorithm}[t]
\caption{The training process of the SCDEM}
\textbf{Input:} Total tasks $N$; Backbones $\{ f_{\theta_1}, \cdots, f_{\theta_{t'}} \}$; Depth $L$ \;
\textbf{Output:} $\{ \mathcal{E}_t \}_{t=1}^{N}$, Updated $\{f_{\theta_j}\}$ \;
\textbf{Init: }\text{Freeze $\{ f_{\theta_j} \}_{j=1}^{t'}$ except for last $L$ layers;}

\For{$t = 1$ \KwTo $N$}{
    Create new expert $\mathcal{E}_t = \{ f_{\xi_t}, f_{\omega_t} \}$ \;
    \lIf{$t > 1$}{Create selectors $\{ g_{\phi_1}, \cdots, g_{\phi_{t'}} \}$}

    \textbf{Training:}
    \For{$\{\mathbf{x}, \mathbf{y}\} \in D^s_t$}{
        ${\bf z}^f = \bigotimes\nolimits_{j=1}^{t'} f_{\theta_j}({\bf x})$, \quad $\hat{\bf y} = f_{\omega_t}(f_{\xi_t}({\bf z}^f))$ \;
        $\textbf{Step 1: }\mathcal{L}_{\rm cls} = \mathcal{L}_{\rm CE}(\hat{\bf y}, {\bf y})$ \;

        \If{$t > 1$}{
            $\textbf{Step 2: } {\hat{\bf z}}^f = \bigotimes\nolimits_{j=1}^{t'} \hat{f}_{\theta_j}({\bf x})$ \;
            $\mathcal{L}_{\rm COM} = \sum\limits_{i=1}^{t-1} D_{\rm KL}\left[ f_{\omega_i}(f_{\xi_i}({\bf z}^f)) \parallel f_{\omega_i}(f_{\xi_i}(\hat{\bf z}^f)) \right]$ \;
            
            $\textbf{Step 3: }\text{Get features }\ \mathcal{Z}_j,\ \hat{\mathcal{Z}_j}$\\
            
            $\boldsymbol{\alpha}_j = \mathrm{Softmax}(g_{\phi_j}([ {\bf z}^{j,1},{\bf z}^{j,2}, \dots, {\bf z}^{j,L} ]))$ \;
            $\hat{\boldsymbol{\alpha}}_j = \mathrm{Softmax}(g_{\phi_j}([ \hat{\bf z}^{j,1},\hat{\bf z}^{j,2}, \dots, \hat{\bf z}^{j,L} ]))$ \;
            
            ${\bf Z}^{\text{fused}}_j = \sum\nolimits \alpha_j[k]\cdot{\bf z}^{j,k},\; \  \hat{\bf Z}^{\text{fused}}_j = \sum\nolimits \hat{\alpha}_j[k]\cdot\hat{\bf z}^{j,k}$ \;
            
            $\mathcal{L}_{\rm Fused} = \sum\nolimits_{j=1}^{t'} \mathcal{W}(P_{{\bf Z}^{\text{fused}}_j}, P_{{\hat{\bf Z}}^{\text{fused}}_j})$ \;
        }

        $\mathcal{L}_{\rm total} = \mathcal{L}_{\rm cls} + \mathcal{L}_{\rm COM} + \mathcal{L}_{\rm Fused}$ \;
        \textbf{Step 4: }Update $\{ f_{\xi_t}, f_{\omega_t}, \theta^{(L)}_j, \phi_t \}$ by $\nabla \mathcal{L}_{\rm total}$ \;
    }

    \textbf{Snapshot: }$\{\hat{f}_{\theta_j}\} \leftarrow \mathrm{copy}\{f_{\theta_j}\}; \quad \text{Freeze } \{\hat{f}_{\theta_j} \}, \  \mathcal{E}_t$\;
}
\end{algorithm}

In continual learning scenarios, existing research often introduces a new, independent expert module in mixture systems to begin training with minimal parameters. This approach can employ a single pre-trained ViT as the backbone network that contains only a small subset of the semantic knowledge from one or a few data domains. As a result, the model exhibits significant limitations when dealing with data from domains that have large distributional shifts. Additionally, the parameters of the backbone in these dynamic expansion models are usually frozen during training, which reduces the model's generalization ability and adaptability to the newly seen data domain. To address these issues, we propose a novel dynamic expansion model, which manages and optimizes several different backbone networks that were trained on different data sources. Such a dynamic expansion model demonstrates strong generalization across various domains while mitigating catastrophic forgetting of previous knowledge. The overall architecture of the proposed framework is shown in ~\cref{fig1}, and the individual network components will be discussed in detail in the following sections.

\vspace{2pt}
\noindent
\textbf{The multi-source backbones.} Utilizing multiple different backbones, each trained on distinct datasets and domains, can produce richer, more versatile feature representations that significantly enhance the model’s capacity in continual learning scenarios. Let $\{ {f_{\theta_1}, \dots, f_{\theta_{t'}}} \}$ represent a collection of $t'$ distinct backbones, where each backbone $f_{\theta_j} \colon \mathcal{X} \to \mathcal{Z}$ is implemented using a pre-trained ViT \cite{Vit}, where an input image ${\bf x} \in \mathcal{X}$ is mapped to a feature vector ${\bf z} \in \mathcal{Z}$, with $\mathcal{Z} \subseteq \mathbb{R}^{d_z}$ representing the feature space of dimension $d_z$, and $\theta_j$ denoting the parameters of the $j$-th backbone. To minimize computational cost while retaining key information, we extract and use only the class token from each backbone’s output as representations. Given an input ${\bf x}$, we can leverage all $t'$ pre-trained backbones to generate a robust feature representation by concatenating their outputs as follows~:
\begin{equation} 
\begin{aligned}
{\bf z}^f = {\bf z}^1 \otimes {\bf z}^2 \otimes \cdots \otimes {\bf z}^{t'}\,, \label{combinedFeatures} 
\end{aligned}
\end{equation} 
where ${\bf z}^j$ represents the feature vector produced by the $j$-th backbone $f_{\theta_j}$, and $\otimes$ indicates the concatenation of these vectors. The resulting feature vector ${\bf z}^f$ lies in an augmented feature space ${\mathcal{Z}}^f \in \mathbb{R}^{d_z \times t'}$.

\vspace{2pt}
\noindent
\textbf{The expert module.} Although pre-trained backbones are effective at producing rich feature representations, they cannot be directly used for making predictions on new tasks. To address this issue, we propose a new creation approach to dynamically construct and integrate an expert module within a flexible expansion framework to learn the decision boundary for a new task. Specifically, for a given task ${T}_j$, we design a new expert module ${\mathcal{E}}_j$, which consists of an adaptive module $f_{\xi_j} \colon \mathcal{Z}^f \to \mathcal{Z}^e$ that learns a task-specific representation, and a linear classifier $f_{\omega_j} \colon \mathcal{Z}^e \to \mathcal{Y}$ that identifies the decision-making pattern for the task. The adaptive module $f_{\xi_j}$ processes the augmented feature vector ${\bf z}^f$ and generates a new feature vector ${\bar{\bf z}}^j$ in the feature space ${\mathcal{Z}}^e \subseteq \mathbb{R}^{d_e}$, where $d_e$ represents the dimensionality of the learned task-specific features. The prediction process using the $j$-th expert for a given data sample ${\bf x}$ is expressed as~: \begin{equation}
\begin{aligned}
y' = \arg\max \left( {\rm Softmax}\left( {\bf W}^{\rm T}{\omega_j} {\bar{\bf z}}^j \right) \right)\,,
\label{classifierEq} 
\end{aligned}
\end{equation} 
where ${\bf W}{\omega_j}$ is the weight matrix of the classifier $f_{\omega_j}$, and ${\rm Softmax}(\cdot)$ denotes the Softmax activation function. ${\bf W}^{\rm T}_{\omega_j}$ represents the transpose of the weight matrix and $y'$ is the predicted class label.

\begin{table*}[t]
\centering
\caption{Performance comparison of SCDEM and SOTA models in a dual-domain task configuration. "Average" denotes mean performance across all tasks, while "Last" shows the performance on the final task. All results are averaged over 10 runs. SCDEM$^2$ or SCDEM$^3$ indicates the use of 2 or 3 backbones respectively.}
\vspace{-8pt}
\fontsize{9pt}{11pt}\selectfont 
\setlength{\tabcolsep}{0.5pt} 
\begin{tabularx}{\textwidth}{l*{12}{>{\centering\arraybackslash}X}} 
\toprule
\textbf{Method} & \multicolumn{2}{c}{\textbf{TinyImage-Birds}} & \multicolumn{2}{c}{\textbf{Birds-TinyImage}} & \multicolumn{2}{c}{\textbf{Cifar10-Birds}} & \multicolumn{2}{c}{\textbf{Birds-Cifar10}} & \multicolumn{2}{c}{\textbf{Cifar100-Birds}} & \multicolumn{2}{c}{\textbf{Birds-Cifar100}} \\
 & Average & Last & Average & Last & Average & Last & Average & Last & Average & Last & Average & Last \\
\midrule

DER \cite{CL_DarkKD} & 83.7\text{\scriptsize ±2.92} & 99.8\text{\scriptsize ±0.44} & 90.1\text{\scriptsize ±1.54} & 93.3\text{\scriptsize ±0.43} & 85.2\text{\scriptsize ±3.60} & 99.8\text{\scriptsize ±0.44} & 98.3\text{\scriptsize ±0.27} & 96.3\text{\scriptsize ±0.31} & 86.9\text{\scriptsize ±2.90} & 99.6\text{\scriptsize ±0.55} & 94.0\text{\scriptsize ±0.76} & 95.3\text{\scriptsize ±0.49} \\

DER++ \cite{CL_DarkKD} & 95.6\text{\scriptsize ±0.41} & 99.5\text{\scriptsize ±0.23} & 95.2\text{\scriptsize ±0.24} & 93.1\text{\scriptsize ±1.00} & 98.7\text{\scriptsize ±0.34} & 99.2\text{\scriptsize ±1.30} & 99.3\text{\scriptsize ±0.06} & 96.1\text{\scriptsize ±0.41} & 97.5\text{\scriptsize ±0.11} & 99.6\text{\scriptsize ±0.89} & 97.1\text{\scriptsize ±0.07} & 95.1\text{\scriptsize ±0.19} \\

DER+++refresh \cite{der+++refresh} & 95.7\text{\scriptsize ±0.36} & 99.5\text{\scriptsize ±0.19} & 95.1\text{\scriptsize ±0.24} & 93.5\text{\scriptsize ±0.47} & 98.7\text{\scriptsize ±0.33} & 99.5\text{\scriptsize ±0.24} & 99.4\text{\scriptsize ±0.09} & 96.2\text{\scriptsize ±0.28} & 97.5\text{\scriptsize ±0.19} & 99.5\text{\scriptsize ±0.23} & 97.3\text{\scriptsize ±0.18} & 95.6\text{\scriptsize ±0.47} \\

MoE-2E/1R \cite{BoostingCL} & 26.7\text{\scriptsize ±0.85} & 76.0\text{\scriptsize ±0.31} & 20.0\text{\scriptsize ±3.63} & 92.1\text{\scriptsize ±1.21} & 26.5\text{\scriptsize ±5.52} & 70.0\text{\scriptsize ±0.54} & 33.9\text{\scriptsize ±0.43} & 96.8\text{\scriptsize ±0.35} & 33.4\text{\scriptsize ±0.52} & 96.2\text{\scriptsize ±0.32} & 37.4\text{\scriptsize ±0.22} & 93.5\text{\scriptsize ±0.82} \\

FDR \cite{benjamin2018measuring} & 21.9\text{\scriptsize ±3.34} & 98.6\text{\scriptsize ±1.51} & 35.3\text{\scriptsize ±7.61} & 92.7\text{\scriptsize ±0.47} & 65.9\text{\scriptsize ±8.78} & 99.0\text{\scriptsize ±1.73} & 70.9\text{\scriptsize ±4.21} & 95.9\text{\scriptsize ±0.06} & 48.9\text{\scriptsize ±12.6} & 99.0\text{\scriptsize ±0.43} & 68.2\text{\scriptsize ±7.29} & 94.9\text{\scriptsize ±0.68} \\

AGEM-R \cite{chaudhry2018efficient} & 39.6\text{\scriptsize ±14.7} & 77.8\text{\scriptsize ±40.7} & 64.9\text{\scriptsize ±10.2} & 92.3\text{\scriptsize ±0.75} & 46.6\text{\scriptsize ±18.1} & 99.2\text{\scriptsize ±0.84} & 95.4\text{\scriptsize ±1.58} & 96.0\text{\scriptsize ±0.21} & 40.6\text{\scriptsize ±8.29} & 98.8\text{\scriptsize ±1.30} & 83.7\text{\scriptsize ±4.07} & 95.3\text{\scriptsize ±0.30}\\

iCaRL \cite{icarl} & 64.8\text{\scriptsize ±1.21} & 32.2\text{\scriptsize ±6.06} & 66.2\text{\scriptsize ±0.62} & 91.8\text{\scriptsize ±0.45} & 12.3\text{\scriptsize ±0.76} & 2.00\text{\scriptsize ±2.34} & 35.2\text{\scriptsize ±2.69} & 96.3\text{\scriptsize ±0.17} & 58.1\text{\scriptsize ±1.49} & 42.8\text{\scriptsize ±5.22} & 55.4\text{\scriptsize ±1.53} & 94.3\text{\scriptsize ±0.14} \\

\rowcolor{lightgray}
StarPrompt \cite{starprompt} & \textbf{97.8\text{\scriptsize ±0.48}} & \textbf{98.9\text{\scriptsize ±0.82}} & \textbf{97.8\text{\scriptsize ±0.79}} & \textbf{96.3\text{\scriptsize ±0.91}} & \textbf{99.2\text{\scriptsize ±0.37}} & \textbf{99.0\text{\scriptsize ±0.71}} & \textbf{99.2\text{\scriptsize ±0.46}} & \textbf{98.0\text{\scriptsize ±1.01}} & \textbf{98.3\text{\scriptsize ±0.59}} & \textbf{96.8\text{\scriptsize ±1.32}} & \textbf{98.2\text{\scriptsize ±0.52}} & \textbf{97.6\text{\scriptsize ±0.92}} \\

RanPac \cite{mcdonnell2023ranpac} & 93.8\text{\scriptsize ±0.88} & 91.2\text{\scriptsize ±0.31} & 94.1\text{\scriptsize ±0.65} & 95.9\text{\scriptsize ±0.43} & 98.9\text{\scriptsize ±0.74} & 93.1\text{\scriptsize ±0.53} & 98.7\text{\scriptsize ±0.92} & 98.7\text{\scriptsize ±0.42} & 95.4\text{\scriptsize ±0.95} & 88.6\text{\scriptsize ±0.32} & 95.4\text{\scriptsize ±0.32} & \textbf{98.6\text{\scriptsize ±0.32}}\\

Dap \cite{dap} & 92.9\text{\scriptsize ±0.72} & 95.0\text{\scriptsize ±0.89} & 92.4\text{\scriptsize ±0.52} & 93.4\text{\scriptsize ±0.41} & 83.4\text{\scriptsize ±0.67} & 97.9\text{\scriptsize ±0.88} & 90.7\text{\scriptsize ±0.42} & 99.0\text{\scriptsize ±0.32} & 90.4\text{\scriptsize ±0.52} & 94.8\text{\scriptsize ±0.42} & 90.6\text{\scriptsize ±0.68} & 98.0\text{\scriptsize ±0.72} \\

\midrule
\rowcolor{lightpink}
SCDEM$^2$(Ours) & \textbf{97.2\text{\scriptsize ±0.08}} & \textbf{99.6\text{\scriptsize ±0.18}} & \textbf{97.0\text{\scriptsize ±0.15}} & \textbf{93.9\text{\scriptsize ±0.44}} & \textbf{99.3\text{\scriptsize ±0.11}} & \textbf{99.7\text{\scriptsize ±0.19}} & \textbf{99.2\text{\scriptsize ±0.18}} & \textbf{96.4\text{\scriptsize ±0.25}} & \textbf{97.8\text{\scriptsize ±0.14}} & \textbf{99.7\text{\scriptsize ±0.12}} & \textbf{97.6\text{\scriptsize ±0.12}} & \textbf{95.4\text{\scriptsize ±0.33}} \\

Rel.ER vs DER+++re & $\downarrow$ 34.88\% & $\downarrow$ 21.56\% & $\downarrow$ 38.77\% & $\downarrow$ 6.15\% & $\downarrow$ 46.92\% & $\downarrow$ 41.17\% & $\uparrow$ 35.11\% & $\downarrow$ 5.26\% & $\downarrow$ 12.4\% & $\downarrow$ 41.27\% & $\downarrow$ 11.44\% & $\uparrow$ 51.22\% \\

SCDEM$^3$(Ours) & \textbf{97.9\text{\scriptsize ±0.74}} & \textbf{99.6\text{\scriptsize ±0.27}} & \textbf{98.0\text{\scriptsize ±0.42}} & \textbf{97.9\text{\scriptsize ±0.41}} & \textbf{99.4\text{\scriptsize ±0.83}} & \textbf{99.2\text{\scriptsize ±0.56}} & \textbf{99.6\text{\scriptsize ±0.56}} & \textbf{98.0\text{\scriptsize ±1.36}} & \textbf{98.4\text{\scriptsize ±0.83}} & \textbf{99.2\text{\scriptsize ±0.55}} & \textbf{98.3\text{\scriptsize ±1.11}} & \textbf{97.2\text{\scriptsize ±0.76}} \\

Rel.ER vs StarPrompt & $\downarrow$ 4.97\% & $\downarrow$ 63.96\% & $\downarrow$ 9.50\% & $\downarrow$ 43.39\% & $\downarrow$ 25.92\% & $\downarrow$ 20.79\% & $\downarrow$ 50.62\% & $\downarrow$ 0\% & $\downarrow$ 6.43\% & $\downarrow$ 12.77\% & $\downarrow$ 6.07\% & $\uparrow$ 17.01\% \\

\bottomrule
\end{tabularx}
\label{tab:shortdomainAcc}
\vspace{-9pt}
\end{table*}

\subsection{Collaborative Optimization Mechanism}

Freezing all backbone networks can mitigate catastrophic forgetting; however, it constrains adaptability in acquiring new tasks due to the limited activation of parameters. To address this challenge and improve the model's generalization capabilities in new task learning, we propose optimizing only a select few of the final $L$ trainable representation layers of each backbone $f_{\theta_j}$, where $j=1,\cdots,t'$. Notably, optimizing these trainable representation layers during new task learning may induce catastrophic forgetting in each historical expert. To counteract this, we introduce an innovative Collaborative Optimization Mechanism (COM) designed to incrementally optimize each backbone while minimizing significant forgetting.

Specifically, before training on a new task ($T_j$), we preserve and freeze the trainable parameters of all backbone networks, denoted as $\{\hat{f}_{\theta_1}, \dots, \hat{f}_{\theta_{t'}}\}$, forming a static historical knowledge framework. Given an input sample ${\bf x} \in \mathcal{X}$, we can get augmented features ${\bf z}^{f}$ and ${\hat{\bf z}}^{{f}}$ extracted from the activated backbones $\{f_{\theta_1},\cdots,f_{\theta_{t'}} \}$ and the frozen historical backbones $\{\hat{f}_{\theta_1}, \dots, \hat{f}_{\theta_{t'}}\}$, respectively. By using ${\bf z}^f$ and ${\hat{\bf z}}^{f}$, each expert ${\mathcal{E}}_j$ can give the task-specific representations, expressed as~:
\begin{equation}
\begin{aligned}
   {\bar{\bf z}}^j = f_{\xi_j} ( {\bf z}^f)\,,  {\hat{\bf z}}^j = f_{\xi_j} ( {\hat{\bf z}}^f) \,.
\end{aligned}
\end{equation}

By utilizing the extracted features, we can create two predictive distributions $p({\bf y}\,|\, {\bar{\bf z}}^i, \mathcal{E}_i)$ and $p({\bf y}\,|\, {\hat{\bf z}}^i, \mathcal{E}_i)$ in which the variable ${\bf y}$ relies on the feature ${\bar{\bf z}}^i$ extracted from the activated and frozen backbones, respectively. As a result, the proposed COM minimizes the probability distance between two predictive distributions, expressed as~:
\begin{equation}
\begin{aligned}
    \mathcal{L}_{\text{COM}} = \sum_{i=1}^{j-1} D_{\text{KL}}\left(p({\bf y} \,|\, {\bar{\bf z}}^i, \mathcal{E}_i) \parallel p({\bf y} \,|\, {\hat{\bf z}}^{i}, \mathcal{E}_i)\right)\,,
    \label{Eq:COM}
\end{aligned}
\end{equation}
where $D_{\rm KL}(\cdot)$ is the Kullback–Leibler (KL) divergence. In practice, each $p({\bf y} \,|\, {\bar{\bf z}}^i, \mathcal{E}_i)$ is implemented using the softmax activate function of the classifier, expressed as $f_{\omega_j}( {\bar{\bf z}}^i )$. As a result, Eq.~\eqref{Eq:COM} can be rewritten as~:
\begin{equation}
\begin{aligned}
    \mathcal{L}'_{\text{COM}} = \sum_{i=1}^{j-1}
    \Big\{
    \sum_{c=1}^{U} \Big\{
f_{\omega_j}( {\bar{\bf z}}^i )[c] \frac{f_{\omega_j}( {\bar{\bf z}}^i )[c]}{f_{\omega_j}( {\Tilde{\bf z}}^i )[c]} \Big\} \Big\} \,,
    \label{Eq:COM2}
\end{aligned}
\end{equation}
\noindent
where $f_{\omega_j}( {\bar{\bf z}}^i )[c]$ denotes the $c$-th dimension of the prediction $f_{\omega_j}( {\bar{\bf z}}^i)$ and $U$ is the total number of classes. Eq.~\eqref{Eq:COM2} can ensure that optimizing the parameters of these backbones does not influence the previously learnt prediction ability of each history expert.

\subsection{Feature Distribution Consistency via Wasserstein Distance}

In addition to ensure that the outputs of the expert modules within the activated backbones $\{f_{\theta_1}, \dots, f_{\theta_{t'}}\}$ are consistent with those of the historical backbones $\{\hat{f}_{\theta_1}, \dots, \hat{f}_{\theta_{t'}}\}$ across all previously encountered tasks, it is imperative to preserve the semantic congruence of the representations derived from both the activated and frozen backbones. This strategy effectively mitigates the adverse effects of negative knowledge transfer. To achieve this objective, we introduce an innovative Feature Distribution Consistency (FDC) method, which quantifies the feature distribution divergence between corresponding layers through the application of the Wasserstein distance \cite{Wasserstein_TFCL}. The Wasserstein distance is based on the transport distance theory and has several advantages~: (1) It provides meaningful gradients even when two target distributions are disjoint; (2) It encourages the generator to cover the entire support of the real data distribution, compared to other distance measures such as KL and JS divergence. Specifically, we define a feature extraction function to derive a layer-specific representation, denoted as:
\begin{equation}
\begin{aligned}
\hspace{-8pt}
    F_{\rm t}(f_{\theta_j},{\bf x},k) = 
    \begin{cases}
     f_{\theta^1_j} ({\bf x})  & k = 1 \\  f_{\theta^2_j} (
 f_{\theta^1_j}  ({\bf x}))
       & k=2 \\
      f_{\theta^k_j} ( \cdots  f_{\theta^1_2} (
 f_{\theta^1_j}  ({\bf x})))
       & 3 \le k \le L\,,
    \end{cases}
    \label{eq:extractFeature}
    \end{aligned}
\end{equation}
where $ f_{\theta^k_j} $ denotes the $k$-th trainable layer of the backbone $f_{\theta_j}$, which receives the feature vector from the $(k-1)$-th trainable layer and returns a representation. By using Eq.~\eqref{eq:extractFeature}, a set of feature vectors extracted by a backbone can be expressed as~:
\begin{equation}
\begin{aligned}
{\bf Z}^{j,k} = \{ {\bf z}_c \,|\, {\bf z}_c = F(f_{\theta_j}, {\bf x}_c, k ), c=1,\cdots,b \}\,, 
\end{aligned}
\end{equation}
where $j=1,\cdots,t'$ and $k=1,\cdots,L$ denote the index of the expert and trainable representation layer, respectively. Let $P_{{\bf Z}^{j,k}}$ denote the probability distribution of ${\bf Z}^{j,k}$. The proposed FDC approach minimizes the Wasserstein distance between distributions~:
\begin{equation}
\begin{aligned}
{\mathcal{L}}_{\rm FDC} = \sum^{t'}_{j=1}\Big\{ \sum^{L}_{k=1}\Big\{ 
{\mathcal{W}}( P_{{\bf Z}^{j,k}} , P_{\hat{\bf Z}^{j,k}} )
\Big\}\Big\}\,,
\label{eq:FDC}
\end{aligned}
\end{equation}
where $P_{\hat{\bf Z}^{j,k}}$ is the distribution of the representations returned using $F_t( {\hat f}_{\theta_j}, {\bf x}, k )$ and ${\mathcal{W}}(\cdot,\cdot)$ denotes the Wasserstein distance.

\subsection{Dynamic Layer-Wise Feature Attention Mechanism}
Different layers within backbone networks capture features at varying semantic granularities. Shallow layers generally encode low-level visual information, whereas deeper layers provide task-specific semantic abstractions. Consequently, each layer contributes differently when adapting to new tasks. To dynamically balance these multi-layer representations, we propose an adaptive feature fusion mechanism using a learnable attention network.

Formally, given the last $L$ trainable representation layers from the backbone $f_{\theta_j}$, we construct the layer-wise feature set as $\mathcal{Z}_j = [{\bf z}^{j,1}, {\bf z}^{j,2}, \dots, {\bf z}^{j,L}]\in\mathbb{R}^{L\times d_z}$, where ${\bf z}^{j,k}$ denotes the feature vector extracted using the $k$-th feature layer of the $j$-th backbone. To dynamically determine each layer's contribution, we introduce a learnable attention network $g_{\phi_t}(\cdot)$ named selector parameterized by $\phi_t$, which jointly processes the entire feature set and outputs a vector of layer-specific logits:
\begin{equation}
\begin{aligned}
{\boldsymbol{\alpha}}_j = \left\{ 
\alpha_k \;\middle|\;
\alpha_k = \frac{\exp\left(g_{\phi_t}({\bf z}^{j,k} )\right)}{\sum\limits_{l=1}^{L} \exp\left(g_{\phi_t}( {\bf z}^{j,l} )\right)} 
\;,\; k = 1, \dots, L 
\right\}
\label{eq:fusion}
\end{aligned}
\end{equation}
where ${\boldsymbol{\alpha}}_j$ denotes the adaptive weight for the trainable representation layers of the $j$-th backbone. By using Eq.~\eqref{eq:fusion}, we can extend the layer-wise features into a single unified representation as~:
\begin{equation}
\begin{aligned}
{\bf Z}^{\text{fused}}_j = \sum_{k=1}^{L}\Big\{  {{\boldsymbol{\alpha}}_j}[k] \cdot{\bf z}^{j,k} \Big\} ,\quad
{\hat{\bf Z}}^{\text{fused}}_j = \sum_{k=1}^{L}
\Big\{{{\boldsymbol{\hat{\alpha}}}_j}[k] \cdot{\hat{\bf z}}^{j,k} \Big\}\,,
\end{aligned}
\end{equation}
\noindent
where $\hat{\boldsymbol{\alpha}}_j[k]$ denotes the adaptive weight of the $k$-th representation layer of the $j$-th frozen backbone $\hat{f}_{\theta_j}$. To enforce semantic consistency and prevent forgetting during incremental learning, we minimize the Wasserstein distance between the distributions of current fused features ${\bf z}^{\text{fused}}_j$ and historical fused features ${\hat{\bf z}}^{\text{fused}}_j$, resulting in~:
\begin{equation}
\begin{aligned}
\mathcal{L}_{\text{Fused}} = \sum_{j=1}^{t'} \mathcal{W}\left(P_{{\bf Z}^{\text{fused}}_j},\,P_{{\hat{\bf Z}}^{\text{fused}}_j}\right)\,,
\label{eq:fused}
\end{aligned}
\end{equation}
\noindent
where $P_{{\bf Z}^{\text{fused}}_j}$ and $P_{{\hat{\bf Z}}^{\text{fused}}_j}$ represent the distributions of fused features from the current and historical backbones, respectively. The parameters of $g_{\phi_t}(\cdot)$ are optimized jointly with backbone parameters during the new task learning, allowing the model to dynamically prioritize informative layers according to task-specific demands. Compared to the regularization loss term defined in Eq.~\eqref{eq:FDC}, Eq.~\eqref{eq:fused} can adaptively penalize the changes on each trainable representation layer of the backbones, which avoids over-regularization issues and reduces computational costs. 

\vspace{-2pt}
\subsection{Algorithm Implementation}

\noindent
The training procedure of SCDEM, summarized in \textbf{Algorithm 1}, consists of four main stages~:

\vspace{2pt}
\noindent
\textbf{Step 1: Supervised classification.} For each task $T_t$, a task-specific expert $\mathcal{E}_t = \{ f_{\xi_t}, f_{\omega_t} \}$ is instantiated. It takes as input the concatenated multi-domain representation ${\bf z}^f$, produced by applying all active backbones $\{ f_{\theta_j} \}$. The prediction $\hat{\bf y}$ is optimized using the cross-entropy loss $\mathcal{L}_{\rm cls}$.

\vspace{2pt}
\noindent
\textbf{Step 2: Collaborative optimization.} To mitigate forgetting, frozen versions of backbones $\{ \hat{f}_{\theta_j} \}$ are preserved before each task. During training, we compute ${\hat{\bf z}}^f$ using the frozen backbones, and constrain the predictive behaviour of all past experts $\{ \mathcal{E}_i \}_{i < t}$ by minimizing the divergence between outputs based on ${\bf z}^f$ and ${\hat{\bf z}}^f$, leading to the distillation loss $\mathcal{L}_{\rm COM}$ by Eq.~\eqref{Eq:COM2}.

\vspace{2pt}
\noindent
\textbf{Step 3: Fused feature consistency.} Instead of constraining each layer individually, we adopt a selector network $g_{\phi_t}$ to assign soft attention weights $\boldsymbol{\alpha}_j$ over the $L$ trainable layers of each backbone as shown in \cref{fig2}(a). These weights are used to generate fused task-aware features ${\bf Z}^{\text{fused}}_j$ and their frozen references ${\hat{\bf Z}}^{\text{fused}}_j$. A Wasserstein-based regularization term $\mathcal{L}_{\rm Fused}$ is introduced to maintain distributional consistency by Eq.~\eqref{eq:fused}, which avoids over-regularization while improving efficiency and robustness.

\vspace{2pt}
\noindent
\textbf{Step 4: Parameter update.} The final loss $\mathcal{L}_{\rm total}$ combines all components above and is used to jointly update the expert $\mathcal{E}_t$, the last $L$ layers of each $f_{\theta_j}$, and the selector $g_{\phi_t}$. After task completion, all backbones are snapshotted and frozen to serve as reference models for future tasks.

\begin{table*}[t]
\centering
\caption{Performance comparison of SCDEM and SOTA in 3-domain and 4-domain configurations, summarizing average performance across all tasks and performance on the final task.}
\vspace{-8pt}
\fontsize{9pt}{11pt}\selectfont 
\setlength{\tabcolsep}{8pt} 
\begin{tabularx}{\textwidth}{l*{8}{>{\centering\arraybackslash}X}} 
\toprule
\textbf{Method} & \multicolumn{2}{c}{\textbf{Tiny-Cifar10-Birds}} & \multicolumn{2}{c}{\textbf{Tiny-Cifar100-Birds}} & \multicolumn{2}{c}{\textbf{Tiny-C100-Birds-C10}} & \multicolumn{2}{c}{\textbf{Average}} \\
 & Average & Last & Average & Last & Average & Last & Average & Last \\
\midrule
DER \cite{CL_DarkKD} & 78.83\text{\scriptsize ±1.07} & 99.65\text{\scriptsize ±0.39} & 66.97\text{\scriptsize ±4.14} & 99.49\text{\scriptsize ±0.22} & 75.42\text{\scriptsize ±3.07} & 96.14\text{\scriptsize ±0.39} & 84.46\text{\scriptsize ±9.51} & 97.78\text{\scriptsize ±2.46}\\

DER++ \cite{CL_DarkKD} & 94.77\text{\scriptsize ±0.20} & 99.50\text{\scriptsize ±0.19} & 93.93\text{\scriptsize ±0.19} & 99.60\text{\scriptsize ±0.89} & 93.73\text{\scriptsize ±0.32} & 95.81\text{\scriptsize ±0.32} & 96.21\text{\scriptsize ±1.95} & 97.62\text{\scriptsize ±2.55}\\

DER+++refresh \cite{der+++refresh} & 94.89\text{\scriptsize ±0.27} & 99.50\text{\scriptsize ±0.12} & 94.26\text{\scriptsize ±0.28} & 99.80\text{\scriptsize ±0.09} & 93.83\text{\scriptsize ±0.30} & 96.45\text{\scriptsize ±0.29} & 96.31\text{\scriptsize ±1.91} & 97.77\text{\scriptsize ±2.41}\\

MoE-2E/1R \cite{BoostingCL} & 31.22\text{\scriptsize ±0.36} & 92.00\text{\scriptsize ±0.41} & 28.83\text{\scriptsize ±0.43} & 91.36\text{\scriptsize ±0.40} & 27.55\text{\scriptsize ±0.51} & 92.33\text{\scriptsize ±0.42} & 29.47\text{\scriptsize ±1.31} & 93.73\text{\scriptsize ±1.79}\\

FDR \cite{benjamin2018measuring} & 24.25\text{\scriptsize ±2.39} & 98.20\text{\scriptsize ±2.05} & 17.28\text{\scriptsize ±1.68} & 98.4\text{\scriptsize ±0.89} & 17.09\text{\scriptsize ±1.75} & 95.20\text{\scriptsize ±0.83} & 41.08\text{\scriptsize ±22.46} & 97.00\text{\scriptsize ±2.50}\\

AGEM-R \cite{chaudhry2018efficient} & 32.71\text{\scriptsize ±2.88} & 74.83\text{\scriptsize ±39.3} & 24.95\text{\scriptsize ±7.86} & 37.87\text{\scriptsize ±45.6} & 47.89\text{\scriptsize ±3.87} & 95.82\text{\scriptsize ±0.43} & 52.94\text{\scriptsize ±24.1} & 85.46\text{\scriptsize ±29.0} \\

iCaRL \cite{icarl} & 49.84\text{\scriptsize ±0.49} & 4.6\text{\scriptsize ±2.61} & 70.59\text{\scriptsize ±1.12} & 41.20\text{\scriptsize ±3.76} & 70.29\text{\scriptsize ±1.64} & 94.60\text{\scriptsize ±0.21} & 53.63\text{\scriptsize ±18.3} & 55.54\text{\scriptsize ±37.6} \\

\rowcolor{lightgray}
StarPrompt \cite{starprompt} & \textbf{97.70\text{\scriptsize ±0.65}} & \textbf{99.12\text{\scriptsize ±0.77}} & \textbf{97.01\text{\scriptsize ±0.15}} & \textbf{98.01\text{\scriptsize ±1.00}} & \textbf{97.39\text{\scriptsize ±0.35}} & \textbf{97.76\text{\scriptsize ±0.71}} & \textbf{98.06\text{\scriptsize ±0.75}} & \textbf{98.14\text{\scriptsize ±0.97}} \\

RanPac \cite{mcdonnell2023ranpac} & 93.92\text{\scriptsize ±0.48} & 91.10\text{\scriptsize ±0.35} & 94.15\text{\scriptsize ±0.38} & 92.32\text{\scriptsize ±0.76} & 94.14\text{\scriptsize ±0.39} & 95.15\text{\scriptsize ±0.60} & 95.38\text{\scriptsize ±2.02} & 93.85\text{\scriptsize ±3.48}\\

Dap \cite{dap} & 94.48\text{\scriptsize ±0.51} & 92.65\text{\scriptsize ±0.45} & 92.77\text{\scriptsize ±0.39} & 95.51\text{\scriptsize ±0.40} & 91.62\text{\scriptsize ±0.47} & 95.83\text{\scriptsize ±0.42} & 91.03\text{\scriptsize ±3.15} & 96.49\text{\scriptsize ±2.15}\\

\midrule
\rowcolor{lightpink}
SCDEM$^2$\text{(Ours)} & \textbf{97.16\text{\scriptsize ±0.06}} & \textbf{99.81\text{\scriptsize ±0.09}} & \textbf{96.43\text{\scriptsize ±0.05}} & \textbf{99.72\text{\scriptsize ±0.13}} & \textbf{96.51\text{\scriptsize ±0.08}} & \textbf{96.6\text{\scriptsize ±0.13}} & \textbf{97.58\text{\scriptsize ±1.05}} & \textbf{98.04\text{\scriptsize ±2.44}}\\

Rel.ER vs DER+++re & $\downarrow$ 44.42\% & $\downarrow$ 64.78\% & $\downarrow$ 37.80\% & $\uparrow$ 38.09\% & $\downarrow$ 43.44\% & $\downarrow$ 4.22\% & $\downarrow$ 34.42\% & $\downarrow$ 12.11\% \\

SCDEM$^3$\text{(Ours)} & 97.83\text{\scriptsize ±0.38} & 99.50\text{\scriptsize ±0.44} & 97.22\text{\scriptsize ±0.41} & 99.42\text{\scriptsize ±0.50} & 97.32\text{\scriptsize ±0.34} & 98.02\text{\scriptsize ±1.38} & 98.28\text{\scriptsize ±0.95} & 98.80\text{\scriptsize ±1.53} \\

Rel.ER vs StarPrompt & $\downarrow$ 5.65\% & $\downarrow$ 4.32\% & $\downarrow$ 7.02\% & $\downarrow$ 70.85\% & $\uparrow$ 2.68\% & $\downarrow$ 11.61\% & $\downarrow$ 11.34\% & $\downarrow$ 35.48\%  \\

\bottomrule
\end{tabularx}
\label{tab:longdomainAcc}
\vspace{-9pt}
\end{table*}

\vspace{-10pt}
\section{Experiment}
\subsection{Experimental Setup}

\noindent
\textbf{Datasets:} The model's performance is evaluated in a continual learning framework across several domains, including CIFAR-10 \cite{CIFAR10}, TinyImageNet \cite{TinyImageNet}, CIFAR-100 \cite{CIFAR10}, and Birds 525 Species.

\noindent
\textbf{Evaluation Metrics:} To evaluate and compare the performance of the model in multi-task scenarios, we employ two key metrics: "Average" and "Last." The "Average" metric computes the mean accuracy across all tasks within a given scenario over all the testing samples, while the "Last" metric focuses on the accuracy achieved on the final task. We provide additional experimental configurations in \textbf{Appendix-B} from SM. 

\vspace{-10pt}
\subsection{Comparison with State-of-the-Art Methods}

In this section, we compare our method with several SOTA continual learning approaches, including experience replay-based methods, dynamic expansion models, and other incremental strategies. For experience replay, we evaluate DER \cite{CL_DarkKD} and its variants DER++ \cite{CL_DarkKD} and DER+++refresh \cite{der+++refresh}, which address catastrophic forgetting by storing and replaying past samples. We also include three feature distillation-based methods: FDR \cite{benjamin2018measuring}, which applies feature regularization; AGEM-R \cite{chaudhry2018efficient}, which adjusts gradients using historical task information; and iCaRL \cite{icarl}, which employs memory and nearest-neighbor classification. All methods are implemented with a dual-ViT backbone, unfreezing the last three layers of each ViT for fine-tuning, and sharing a uniform replay buffer size of 5120. We further compare against Mixture-of-Experts (MoE) models \cite{BoostingCL}, which dynamically activate subsets of experts per task, and incremental learning methods that do not use replay, such as Random Packing (RanPac) \cite{mcdonnell2023ranpac} and Data Augmentation Prompt (Dap) \cite{dap}. Additionally, we also consider employing the prompt-based learning models such as the StarPrompt \cite{starprompt} as another baseline in our comparison, which maintains a balance between new and previous tasks through prompt injection and generated replay. 

\noindent
\textbf{Multi-domain Task Incremental Learning.} We examine a variety of domain combinations, including six two-domain configurations, two three-domain setups, and one four-domain scenario. The performance is evaluated using two key metrics: "Average" and "Last." For the two-domain scenarios, we test different orderings of domains to assess how the models generalize under various configurations. Additionally, we investigate both dual-ViT and triple-ViT approaches to determine whether the inclusion of multiple pre-trained backbones can enhance generalization performance.

\begin{table}[t]
\caption{Comparison of Class-IL accuracy in TinyImageNet.}
\vspace{-8pt}
\centering
\fontsize{8.5pt}{9.5pt}\selectfont
\setlength{\tabcolsep}{2pt}
\renewcommand{\arraystretch}{1.0}
\begin{tabularx}{\columnwidth}{l*{6}{>{\centering\arraybackslash}X}}
\toprule
\textbf{Method} & \multicolumn{2}{c}{\textbf{5 step}} & \multicolumn{2}{c}{\textbf{10 step}} & \multicolumn{2}{c}{\textbf{20 step}} \\
\cmidrule(r){2-3} \cmidrule(r){4-5} \cmidrule(r){6-7}
& \textbf{Avg.} & \textbf{Last} & \textbf{Avg.} & \textbf{Last} & \textbf{Avg.} & \textbf{Last} \\
\midrule
DER \cite{CL_DarkKD}              & 53.89 & 89.25 & 44.41 & 94.40 & 33.26 & 94.81 \\
DER++ \cite{CL_DarkKD}           & 70.12 & 90.20 & 70.61 & 93.45 & 70.92 & 96.20 \\
DER+++refresh\cite{der+++refresh}    & 70.13 & 90.26 & 69.77 & 93.20 & 72.11 & 94.88 \\
MoE-2E/1R \cite{BoostingCL}        & 22.91 & 84.55 & 13.80 & 89.45 &  6.38 & 69.80 \\
iCaRL \cite{icarl}            & 75.08 & 63.75 & 69.56 & 53.55 & 63.03 & 38.80 \\
FDR \cite{benjamin2018measuring}              & 21.01 & 67.02 &  9.56 & 92.90 &  5.36 & 95.60 \\
AGEM-R \cite{chaudhry2018efficient}           & 24.82 & 89.85 & 10.25 & 93.41 &  5.17 & 95.00 \\
RanPac \cite{mcdonnell2023ranpac}           & 72.81 & 69.00 & 72.89 & 70.70 & 73.99 & 74.45 \\
Dap \cite{dap}              & 76.42 & 72.89 & 65.98 & 66.30 & 47.26 & 49.40 \\
StarPrompt \cite{starprompt}       & 87.99 & 86.10 & 86.92 & 85.39 & 86.31 & 85.60 \\
\midrule
SCDEM$^2$(Ours)  & \textbf{92.48} & \textbf{90.20} & \textbf{94.02} & \textbf{92.00} & \textbf{92.73} & \textbf{95.39} \\
\bottomrule
\label{tab:class-il-tiny}
\end{tabularx}
\vspace{-25pt}
\end{table}

\noindent
\textbf{Results Analysis.} The classification performance of our approach, compared with several SOTA methods, is shown in \cref{tab:shortdomainAcc} and \cref{tab:longdomainAcc}. The results clearly indicate that our method, referred to as "Ours," achieves superior average performance in nearly all task configurations when using the dual-ViT setup. In addition, memory replay-based approaches, such as DER, and mixture-of-experts models like MoE tend to show weaker performance in the multi-domain task settings. These methods exhibit strong performance on the current task, reflected in their relatively high “Last” scores, demonstrating limited ability to prevent catastrophic forgetting.

In the dual-domain setting, our method (SCDEM$^3$) outperforms StarPrompt by $17.25\%$ on the Average metric and $20.65\%$ on the Last metric. For the three-domain and four-domain configurations, our method (SCDEM$^3$) shows improvements of $3.33\%$ and $28.93\%$, respectively. It is noteworthy that the performance of the three-backbone network model consistently surpasses that of the dual-backbone network across all task configurations. This suggests that incorporating an additional suitable backbone can further enhance the model's performance.

\begin{table}[t]
\caption{Comparison of Class-IL accuracy in CIFAR100.}
\vspace{-8pt}
\centering
\fontsize{8.5pt}{9.5pt}\selectfont
\setlength{\tabcolsep}{2pt}
\renewcommand{\arraystretch}{1.0}
\begin{tabularx}{\columnwidth}{l*{6}{>{\centering\arraybackslash}X}}
\toprule
\textbf{Method} & \multicolumn{2}{c}{\textbf{5 step}} & \multicolumn{2}{c}{\textbf{10 step}} & \multicolumn{2}{c}{\textbf{20 step}} \\
\cmidrule(r){2-3} \cmidrule(r){4-5} \cmidrule(r){6-7}
& \textbf{Avg.} & \textbf{Last} & \textbf{Avg.} & \textbf{Last} & \textbf{Avg.} & \textbf{Last} \\
\midrule
DER \cite{CL_DarkKD}              & 55.54 & 95.32 & 37.45 & 97.30 & 10.02 & 99.40 \\
DER++ \cite{CL_DarkKD}            & 77.80 & 93.95 & 75.48 & 97.70 & 74.83 & 97.65 \\
DER+++refresh\cite{der+++refresh}    & 77.54 & 95.15 & 76.11 & 96.40 & 76.35 & 97.82 \\
MoE-2E/1R \cite{BoostingCL}        & 85.83 & 89.40 & 84.52 & 87.20 & 84.30 & 83.60 \\
iCaRL \cite{icarl}            & 78.89 & 81.85 & 79.33 & 76.13 & 79.64 & 76.25 \\
FDR \cite{benjamin2018measuring}              & 22.03 & 95.85 & 11.79 & 97.65 &  6.96 & 99.45 \\
AGEM-R \cite{chaudhry2018efficient}           & 22.41 & 95.30 & 14.21 & 98.30 &  7.76 & 98.20 \\
RanPac \cite{mcdonnell2023ranpac}           & 76.85 & 78.65 & 77.03 & 77.20 & 77.12 & 74.20 \\
Dap \cite{dap}              & 40.03 & 38.55 & 24.68 &  5.60 & 13.21 &  0.80 \\
StarPrompt \cite{starprompt}       & 88.32 & 90.45 & 93.62 & 99.00 & 86.16 & 83.80 \\
\midrule
SCDEM$^2$(Ours)  & \textbf{94.61} & \textbf{96.39} & \textbf{96.61} & \textbf{98.20} & \textbf{98.00} & \textbf{98.40} \\
\bottomrule
\label{tab:class-il-cifar100}
\end{tabularx}
\vspace{-25pt}
\end{table}

\noindent
\textbf{Class Incremental Learning.} To accommodate the expert mechanism, our model requires the task identifier during inference, which is typical for the Task-IL scenario. To extend the model's applicability to the Class-IL scenario, we propose a novel approach. Specifically, when a new task is introduced, the fused feature representation from the backbone network is input into all experts, each generating their respective log-likelihoods. By computing the entropy of each expert's distribution and the Kullback-Leibler (KL) divergence between their distributions and the overall fused distribution, we derive a "confidence score" for each expert. The expert with the highest confidence score is selected as the output head. This procedure, illustrated in ~\cref{fig2}(b), does not require the task identifier and involves minimal computational overhead, making it a lightweight expert selection mechanism. Experimental results, summarized in  \cref{tab:class-il-tiny} and  \cref{tab:class-il-cifar100}, show that our model consistently outperforms all other methods across all task configurations, demonstrating its effectiveness and stability in continual learning. We provide additional results in \textbf{Appendix-B} from SM.

\noindent
\textbf{Computational Cost.} We evaluate the computational efficiency of our method in comparison with other baseline approaches by analyzing both computational costs and the number of parameters. A detailed comparison of various models is presented in \cref{tab:performanceCost}, including metrics such as training parameters (M), average GPU memory usage (MiB), and runtime efficiency (it/s). Our proposed framework, which leverages a dual-backbone architecture, demonstrates a clear advantage over existing state-of-the-art (SOTA) methods. When compared to the prominent SOTA method StarPrompt, our approach achieves a reduction in training parameters by $51.08\%$, GPU memory usage by $47.29\%$, and training time by $83.21\%$. These results highlight the efficiency of our method in enhancing continual learning for ViT-based models.

\begin{figure}[t]
  \centering
  \includegraphics[page=3, width=\columnwidth, trim=325 180 340 120, clip]{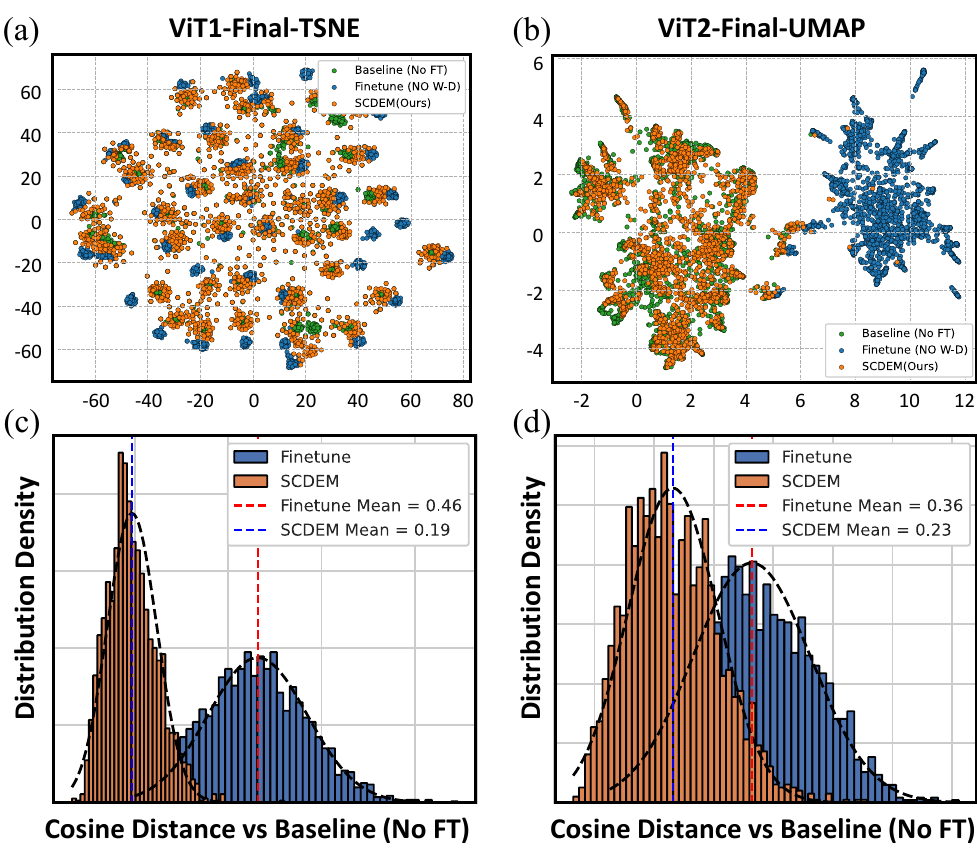}
  \captionsetup{skip=4pt} 
  \caption{(a) and (b) illustrate the feature distributions of the final layer from the dual-backbone network using t-SNE and UMAP, respectively. (c) and (d) compare the cosine distance statistics between the output features and the baseline.}
  \label{fig:feats_map}
  \vspace{-15pt}
\end{figure}

\vspace{-5pt}
\subsection{Ablation study}

\noindent
\textbf{Analysis of Modules.} Table~\ref{tab:ablation_singlecol} reports an ablation study evaluating the contribution of each module in SCDEM$^{2}$ on Tiny-ImageNet and CIFAR100. Removing the COM module (\textit{--No Collaboration}) leads to a performance drop of 1.01\% and 0.42\%, respectively, confirming that dual-backbone representations are not merely additive but synergistic—supporting the learning of richer and more disentangled abstractions. Excluding the Wasserstein Distance constraint (FDC) (\textit{--No W-D Constraint}) yields a drop of 0.45\% and 0.20\%, suggesting that aligning feature distributions across tasks serves as an implicit regularizer, enhancing temporal consistency without explicit memory replay. Removing feature attention (DLWFAM)(\textit{--No Attention}) further reduces accuracy by 0.37\% and 0.14\%, demonstrating its effectiveness in amplifying transferable knowledge while suppressing task-specific noise. Overall, these results underscore that SCDEM$^{2}$ is not a collection of heuristics but a purposefully structured system that balances stability and plasticity through architectural alignment and semantic selection.



\noindent
\textbf{Analysis of W-D Constraint (FDC).}  
To evaluate the impact of feature alignment using Wasserstein Distance (W-D), we randomly selected 20 classes from TinyImageNet and visualized feature distributions via t-SNE and UMAP. As illustrated in \cref{fig:feats_map}.(a) and (b), features constrained by W-D remain significantly closer to the Baseline distribution—i.e., the output of the frozen pretrained backbone—compared to the unconstrained fine-tuned version. This suggests that W-D helps preserve the semantic geometry of the original feature space while enabling adaptation to new tasks.

The histograms in (c) and (d) further quantify this effect: SCDEM achieves notably lower cosine distances to the Baseline (0.23 vs. 0.36 and 0.19 vs. 0.46), reflecting a smaller deviation from the pretrained representations. From a modeling perspective, the fused features are softly regularized toward their historical counterparts, encouraging geometric alignment in both global and local structure. This alignment acts as a structural prior that supports stable yet flexible representation learning across tasks. Additional results are provided in \textbf{Appendix-C} from SM.

\begin{table}[t]
\centering
\caption{Performance comparison from ablation studies on Tiny-ImagNet and Cifar100 by divided into 10 tasks, all results are averaged over 5 runs.}
\vspace{-6pt}
\fontsize{8.0pt}{10pt}\selectfont
\setlength{\tabcolsep}{2pt} 
\renewcommand{\arraystretch}{1.0}
\begin{tabularx}{\linewidth}{l*{4}{>{\centering\arraybackslash}X}}
\toprule
\textbf{Method / Backbone(s)} & \multicolumn{2}{c}{\textbf{Tiny-ImageNet}} & \multicolumn{2}{c}{\textbf{Cifar100}} \\
 & \textbf{Avg.} & \textbf{$\Delta$} & \textbf{Avg.} & \textbf{$\Delta$} \\
 
\midrule
In21k-ft-In1k (ViT\_1) & 94.55 &  0.39$\downarrow$ & 92.62 &  0.39$\downarrow$ \\
In21k (ViT\_2) & 89.03 &  5.91$\downarrow$ & 87.82 &  6.41$\downarrow$ \\
ViT\_1 + ViT\_2 & 93.91 &  1.03$\downarrow$ & 92.48 &  1.75$\downarrow$ \\

\textbf{SCDEM$^{2}$} & \cellcolor{lightpink}\textbf{94.94} & \cellcolor{lightpink}\textbf{--} & \cellcolor{lightpink}\textbf{94.23} & \cellcolor{lightpink}\textbf{--} \\
\phantom{} --No Collaboratio (COM) & 93.93 &  1.01$\downarrow$ & 93.81 &  0.42$\downarrow$ \\
\phantom{} --No W-D Constraint (FDC) & 94.49 &  0.45$\downarrow$ & 94.03 &  0.20$\downarrow$ \\
\phantom{} --No Attention (DLWFAM) & 94.57 &  0.37$\downarrow$ & 94.09 &  0.14$\downarrow$ \\

\bottomrule
\end{tabularx}
\label{tab:ablation_singlecol}
\vspace{-10pt}
\end{table}

\begin{table}[t]
\centering
\caption{Comparison of our method with other SOTA methods in terms of training parameters, GPU usage, and training time. All results are from the "Tiny-Birds" task scenario on RTX 4090 (24GB) and averaged over 5 runs.}
\vspace{-6pt}
\fontsize{8.2pt}{9pt}\selectfont
\setlength{\tabcolsep}{1pt}
\renewcommand{\arraystretch}{1.1}
\begin{tabularx}{\columnwidth}{l*{5}{>{\centering\arraybackslash}X}}
\toprule
\textbf{Method} & \textbf{Params} $\downarrow$ & \textbf{GPU Avg} $\downarrow$ & \textbf{Iteration $\uparrow$} & \textbf{Task Time $\downarrow$} \\
\midrule
DER++ \cite{CL_DarkKD} & 42.27M & 3490 MiB & 3.22 it/s & 110.5s \\
DER+++re \cite{der+++refresh} & 42.27M & 9914 MiB & 2.27 it/s & 357.74s \\
MoE-22E \cite{BoostingCL} & 64.05M & 21362 MiB & 1.93 it/s & 266.65s \\
\rowcolor{lightgray}
StarPrompt \cite{starprompt} & 86.41M & 10112 MiB & 2.49 it/s & 424.19s \\
RanPac \cite{mcdonnell2023ranpac} & 1.49M & 3566 MiB & 3.44 it/s & 250.82s \\
Dap \cite{dap} & 0.68M & 4420 MiB & 2.33 it/s & 147.08s \\
\midrule
SCDEM$^2$\text{(Ours)} & 42.27M & 5330 MiB & 4.71 it/s & 71.05s \\
vs StarPrompt & {\textbf{-51.08\% $\downarrow$}} & {\textbf{-47.29\% $\downarrow$}} & {\textbf{+89.16\% $\uparrow$}} & {\textbf{-83.21\% $\downarrow$}} \\
\bottomrule
\end{tabularx}
\label{tab:performanceCost}
\vspace{-15pt}
\end{table}

\section{Conclusion}
This paper proposes the SCDEM to deal with multiple data domains over time, which can balance adaptability and stability without relying on replay buffers. The three mechanisms, including COM, FDC and DLWFAM are introduced to enhance the adaptability while preventing network forgetting. The empirical results demonstrate that the proposed approach achieves state-of-the-art performance.

{
    \small
    \bibliographystyle{ACM-Reference-Format}
    \bibliography{VAEGAN}
}






\end{document}